\documentclass[letterpaper]{article} 
\usepackage{aaai23}  
\usepackage{times}  
\usepackage{helvet}  
\usepackage{courier}  
\usepackage[hyphens]{url}  
\usepackage{graphicx} 
\usepackage{natbib}  
\usepackage{caption} 
\usepackage{algorithm}

\usepackage{amsmath}
\usepackage{amssymb}
\usepackage{booktabs}
\usepackage{threeparttable}
\usepackage{multirow}
\usepackage{makecell}
\usepackage{longtable}
\usepackage{float}
\usepackage{algpseudocode}

\usepackage{newfloat}
\usepackage{listings}
\DeclareCaptionStyle{ruled}{labelfont=normalfont,labelsep=colon,strut=off} 
\floatstyle{ruled}
\newfloat{listing}{tb}{lst}{}
\floatname{listing}{Listing}

\graphicspath{ {img/} }
\title{X-Transfer: A Transfer Learning-Based Framework for GAN-Generated Fake Image Detection}
\author{
    Lei Zhang${^{1,*}}$, 
    Hao Chen$^{1,}$\thanks{Equal contribution},
    Shu Hu${^2}$,
    Bin Zhu${^3}$,
    Ching Sheng Lin${^4}$,
    Xi Wu${^1}$,
    Jinrong Hu${^{1,\dagger}}$,
    Xin Wang$^{5,}$\thanks{Corresponding author}
}
\affiliations{
    \textsuperscript{\rm 1} Chengdu University of Information Technology\\
    \textsuperscript{\rm 2}  Indiana University–Purdue University Indianapolis \\
    \textsuperscript{\rm 3} Microsoft Research Asia \\
    \textsuperscript{\rm 4} Tunghai University \\
    \textsuperscript{\rm 5} University at Albany, State University of New York (SUNY)

}

\usepackage{bibentry}

\begin{document}

\maketitle

\begin{abstract}
\label{sc:abs}
 Generative adversarial networks ({GANs}) have remarkably advanced in diverse domains, especially image generation and editing. However, the misuse of GANs for generating deceptive images, such as face replacement, raises significant security concerns, which have gained widespread attention. Therefore, it is urgent to develop effective detection methods to distinguish between real and fake images. Current research centers around the application of transfer learning. Nevertheless, it encounters challenges such as knowledge forgetting from the original dataset and inadequate performance when dealing with imbalanced data during training. To alleviate this issue, this paper introduces a novel GAN-generated image detection algorithm called X-Transfer, which enhances transfer learning by utilizing two neural networks that employ interleaved parallel gradient transmission. In addition, we combine AUC loss and cross-entropy loss to improve the model's performance. We carry out comprehensive experiments on multiple facial image datasets. The results show that our model outperforms the general transferring approach, and the best metric achieves 99.04\%, which is increased by approximately 10\%. Furthermore, we demonstrate excellent performance on non-face datasets, validating its generality and broader application prospects. 
\end{abstract}

\section{Introduction}
\label{sc:introd}

\begin{figure}[t]
    \centering
    \includegraphics[width=1.0\linewidth]{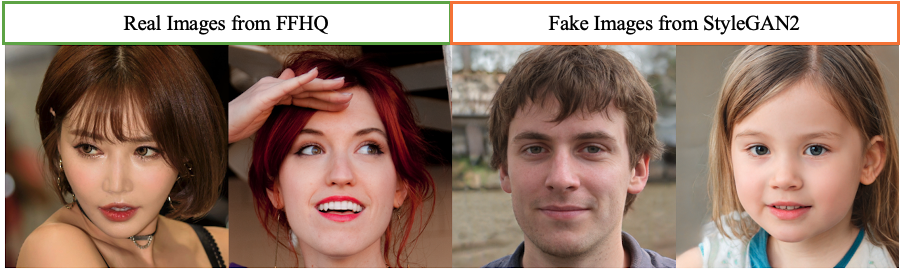}
\caption{Two real images from FFHQ~\cite{karras2019style} (left) and two fake images from StyleGAN2~\cite{karras2020analyzing} (right).}
    \label{fig:fake_imgs}
\end{figure}

Generative adversarial networks (GANs)~\cite{2014Generative} are powerful tools for image generation in computer vision. They can produce highly realistic images that are often indistinguishable from real ones, as shown in Fig.~\ref{fig:fake_imgs}. However, this also raises a serious issue as GAN-generated images can be misused to create fake identities and conduct various malicious activities. Many tools are available for generating fake images, making it effortless to propagate false information. For example, DeepNude~\cite{2021deepfakeabuse} is an open-source deepfake software, initially introduced in 2017 and launched as a website in 2019, that allows users to swap faces and alter body features of images. Such software can be exploited to violate privacy, spread misinformation, and harm reputation. Therefore, it is crucial to develop effective methods to detect fake images generated by GANs and prevent their abuse~\cite{guo2022eyes}. However, existing detection methods still fall short of the expected performance. Detection of GAN-generated images remains a complex and challenging task~\cite{wang2022gan}.



In media forensics, traditional methods~\cite{2019_Detecting,2020Identification} rely on manual statistical features and general classifiers for classification. They use logical-mathematical reasoning to utilize pixels, channels, and other image features for modeling, but they are often tailored to specific training sets and have limited detection accuracy and poor generalization. With the advancement of convolutional neural networks (CNNs)~\cite{2012ImageNet, 2014Very} in feature extraction and recognition, more researchers have adopted CNN-based methods for this task. Traditional methods have thus become less prominent in this field. Many recent studies use CNNs to detect GAN-generated images. Some achieve high detection performance, which can be used to analyze the artifacts or features of GANs~\cite{chen2021locally, do2018forensics, mo2018fake, yang2019exposing}.


Some methods show impressive performance in detecting GAN-generated images, but only on the same dataset they were trained on. They often fail to generalize to new and different datasets. For instance, Bayar's model~\cite{bayar2016deep} exhibits strong performance in detecting image manipulations like median filtering, interpolation, and Gaussian filtering by restricting the initial convolutional layer of the CNNs. However, its performance drops when applied to other GAN domains. This indicates its overfitting to the source dataset and its difficulty in adapting to the target dataset. To overcome this problem, previous studies have used transfer learning~\cite{cozzolino2018forensictransfer,jeon2020t,li2018can}. It is a machine learning technique that adapts a model trained on one task for another related task. It leverages knowledge learned from the source task to enhance performance on the target task, which is especially useful when data for the target task is scarce. The pre-trained model is fine-tuned for the specific needs of the target task. However, applying transfer learning methods directly often causes inconsistencies in performance metrics between datasets. A model may perform well before transfer but poorly after transfer, or vice versa. These variations make it hard to assess the effectiveness of transfer learning methods for detecting GAN-generated images.


In this paper, we aim to achieve high performance on both the source and target datasets during transfer learning without severe knowledge forgetting. Inspired by the {Hybrid Gradient Forward}~\cite{song2021robust} and T-GD~\cite{jeon2020t} methods, we devise a sibling structure for transfer learning. It consists of a master network that learns a new dataset with the assistance of an auxiliary network, which helps preserve the knowledge from the source dataset. The auxiliary network maintains its sibling's familiarity with the original dataset and facilitates its adaptation to new information in the transfer phase. This enables the transfer of essential knowledge from the source to the target dataset, thereby minimizing the loss of prior knowledge. The main contributions of this paper are as follows:


\begin{itemize}
     \item We propose a novel framework, called \textit{X-Transfer}, that includes two sibling networks for transfer learning tasks. It ensures consistent performance before and after transfer learning by balancing hybrid gradient passes and network iterations.
    \item We also adopt the WMW-AUC loss to improve the detection performance. We demonstrate that using the AUC loss enhances the model's generalization ability for detection.
\end{itemize}

This paper is organized as follows. Section~\ref{related} reviews the related works on GAN-generated image detection. Section~\ref{sc:methods} describes our GAN-generated image detection method, including X-Transfer and AUC Loss. Section~\ref{sc:ex} reports experimental results. Section~\ref{sc:conc} concludes the paper.

\section{Related Work}
\label{related}

\subsection{Detection of GAN-generated Images}

With the powerful capabilities of Convolutional Neural Networks (CNNs), the most recent detection methods utilize deep learning methods \cite{guo2021robust}. Mo et al.~\cite{2018Fake} proposed the first CNN-based detection method that focuses on high-frequency components to detect face images generated by Progressive Growing GAN (PGGAN) \cite{karras2017progressive}, achieving impressive performance. They applied a high-pass filter to preprocess an image before inputting it into the CNN, further improving the detection accuracy. However, this method is fragile when faced with image compression and blurs. To address this issue, Nataraj et al.~\cite{2019Detecting} proposed a method that combines co-occurrence matrices and JPEG compression as a tactic for data augmentation. Co-occurrence matrices are extracted from the RGB channels of the input image, then stacked according to the channel dimensions, and finally input into a shallow CNN (such as ResNet-18) to obtain the final detection result. In addition to co-occurrence matrices, researchers have considered introducing preprocessing methods from other traditional forensic techniques into GAN-generated image detection. Mi et al.~\cite{2020GAN} introduced a detection approach employing the self-attention mechanism. This mechanism effectively expands the perception field of convolutional neural networks, enhancing their ability to extract and represent global information. In addition to extracting multiple up-sampling information from GAN more efficiently, it alleviates the spatial distance problem encountered by traditional convolutional neural networks. Their experimental results demonstrate that this method can proficiently identify images generated by PGGAN and exhibits robustness against common post-processing operations.

\begin{figure*}[t]
    \centering
    \includegraphics[scale=0.215]{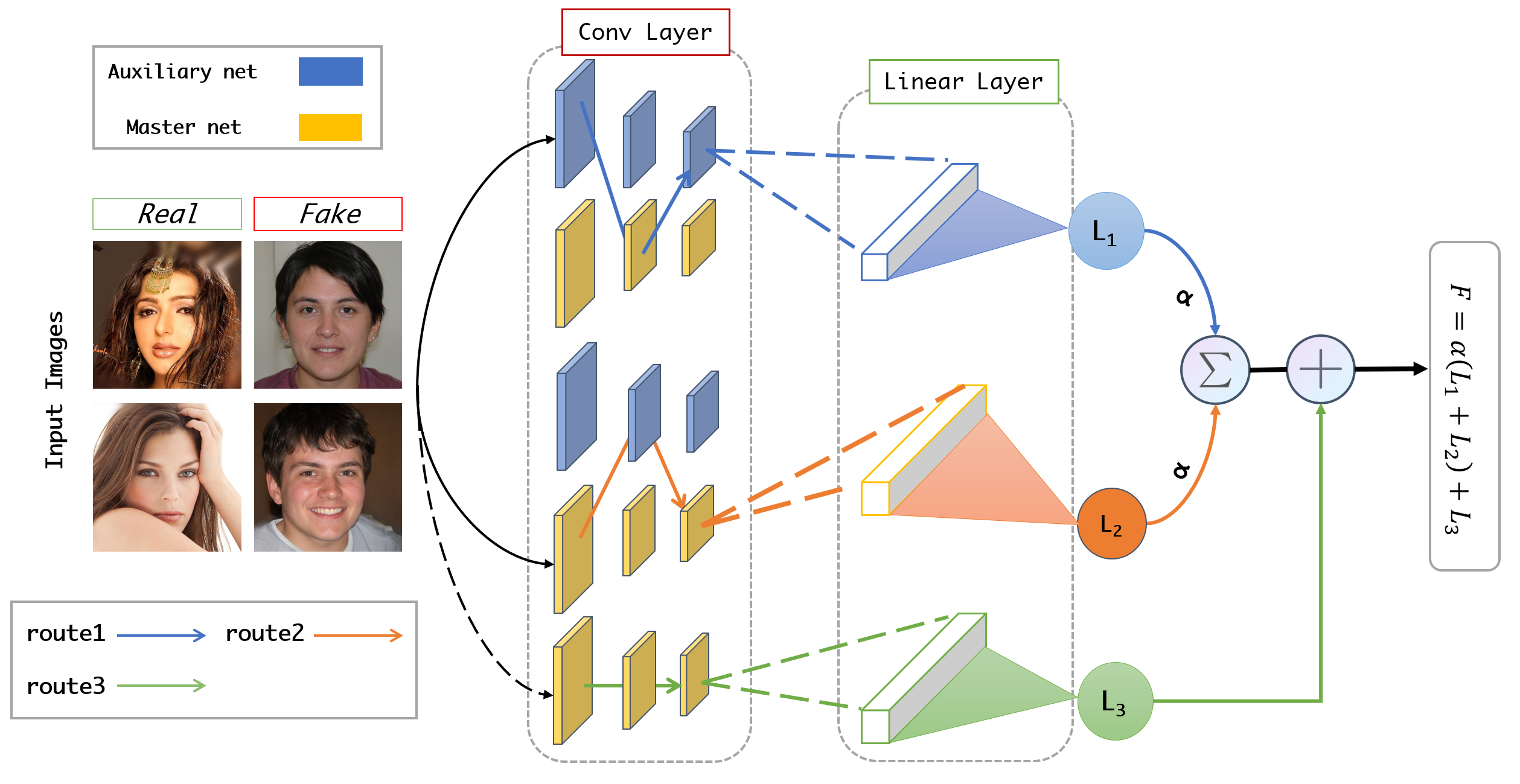}
    \caption{The Structure of the X-Transfer algorithm is depicted as follows. It comprises two identical networks—an auxiliary network and a master network. During the transfer learning process, three routes have been performed in parallel. The $route1$ and $route2$ pass the transferring data through a layer of the
    master network and then a layer of the auxiliary network
    alternately. The difference between them is the former starts with an auxiliary network, and the latter begins with the master network. The loss is denoted as $L_1$ and $L_2$, respectively. For the $L_3$, the data exclusively passes through the master network in $route3$. The final loss function $F=\alpha(L_1+L_2)+L_3$ is a combination of three losses with the weight $\alpha$.}
    \label{fig:architect}
\end{figure*}

\subsection{Data Augmentation for post-processing operation}

One of the challenges in practical applications of image detection generated by GAN is the interference of post-processing operations.
Post-processing operations refer to a series of image or data manipulation steps applied after an initial process, such as resizing, scaling, compression, and filtering. These post-processing operations may degrade the recognition performance of the model.
Wang et al.~\cite{2020CNN} introduced data augmentation methods into their GAN-generated image detection to make a detection method robust to post-processing operations. The authors used ResNet-50~\cite{2016Deep} as the backbone and PGGAN-generated images as the training set. They considered five different data augmentation methods: no data augmentation, Gaussian blur, JPEG compression, Gaussian blur, and JPEG compression with a 50\% probability, and Gaussian blur and JPEG compression with a 10\% probability. Before inputting into the model, samples were cropped and randomly flipped. Their results indicate that data augmentation can enhance the robustness of GAN-generated image detection models when faced with post-processing operations. It can also increase the diversity of training samples and improve detection performance.

In addition to commonly used random rotation, researchers utilize JPEG compression and Gaussian blur. We used CutMix\cite{yun2019cutmix}, random JPEG compression, and random Gaussian kernel in our method.

\subsection{Transfer Learning}

Transfer learning has been widely used in GAN-generated image detection. Wang et al.~\cite{2020CNN} applied ResNet-50 pre-trained on ImageNet as a start, then transferred to their training set. The pre-trained model can enhance the robustness of their model. Hyeonseong et al.~\cite{jeon2020t} employed a neural network based on a teacher-student model, transfer learning, and regularization function. They used a pre-trained network as the teacher model and a starting reference for the student through an $L_2$ regularization function to restrict the variation of convolutional layers between the teacher and student networks, minimizing excessive knowledge forgetting during transfer learning. 

Our method differs in utilizing a mixed gradient propagation approach between two networks for each of their corresponding sub-layers. This approach assists in preventing excessive knowledge forgetting and facilitates knowledge acquisition.

\section{Methodology}
\label{sc:methods}

\subsection{Overview}
We aim to address the issue of declining performance observed in the original model during the transfer learning process. Overall, the architectural framework of our algorithm is illustrated in Figure \ref{fig:architect}. It consists of two integral components: a master network and an auxiliary network, both originating from the same model and being identical to each other. However, they undergo separate routes for updating weights during the transfer process. The input data propagates through alternating sub-networks of the auxiliary network and the master network. Specifically, the forward pass transitions to the other network just before reaching the downsampling layer. This approach is based on the convention that feature maps preceding the downsampling layer are often valuable for task-specific predictions. In other words, the feature map extracted immediately before downsampling provides a robust representation of the image scale. Utilizing an alternate method helps prevent overfitting when assimilating knowledge during the transfer process.

In detail, we define three distinct learning routes stemming from the two networks, master and auxiliary. The first route, starting from the auxiliary network and passing through the master network alternatively, is referred to as ``\textbf{route1}", and its corresponding loss is labeled as $L_1$. Likewise, the second route, emanating from the master network, is named ``\textbf{route2}", with its associated loss indicated as $L_2$. Additionally, the output directly produced by the master model is represented as $L_3$, and its route is termed ``\textbf{route3}." Our primary objective is to achieve a balance among these losses while preserving the prior knowledge acquired during the pre-training phase. For this purpose, we introduce the ``X-Transfer" algorithm, which will be described in Section \textit{X-Transfer}.

To evaluate our approach, we consider that the imbalanced training data would cause the typical evaluation metrics (e.g., accuracy) to be biased by most class samples. We adopt the \textbf{Wilcoxon-Mann-Whitney (WMW)} statistic to simulate the AUC loss function in combination with the binary cross-entropy loss. Through experimental verification, we find that incorporating the AUC Loss yields improved model performance. This highlights the effectiveness of the AUC Loss in training a model that can handle imbalanced datasets more effectively and achieve higher AUC scores.

\subsection{X-Transfer}
\label{sc:x-tr}
Our proposed method is named ``X-Transfer,"  which is phonetically similar to ``Crossed Transfer." As previously mentioned, it primarily comprises two neural networks, a master and an auxiliary network, originating from the same source model. Three distinct routes will be executed throughout the training process and explained as follows.

\textbf{{Route 1}}. This route starts with the auxiliary network; the feed-forward passes the training data through a layer of the master network and then a layer of the auxiliary network alternately, while the backpropagation only updates the layers involved in the feed-forward process and leaves the others fixed. We denoted the loss as $L_1$.

\textbf{{Route 2}}. Likewise with \textit{Route1}, except the difference is that this route originates from the master network. During the feed-forward phase, the training data is sent through sub-networks alternatively within both the master and auxiliary networks. However, updates are exclusively applied to the layers participating in the feed-forward process during backpropagation, while the remaining layers are unchanged. We've designated the resulting loss as $L_2$.

\textbf{{Route 3}}. Differing from the previous routes, this particular route distinguishes itself in terms of loss propagation. It commences from the master neural network only. It propagates data directly through all the layers of the master network sequentially without branching into distinct sub-layers of the auxiliary network. We refer to this cross-entropy loss as $L_3$.

Our target is to balance the multi-losses above three routes and finally retrieve the transferred model of the master network. Let $\{ (x_i, y_i) \}_{i=1}^M$ be a mini-batch of target dataset where $x_i$ represents the $i$th data and $y_i$ is the true label for $x_i$, $M$ stands for the size of mini-batch. In a mini-batch,  $L_1, L_2, L_3$ are the binary cross entropy (BCE) losses. Each of them can be calculated by $\frac{1}{M} \sum_{i=1}^{M} J(y_i, \hat{y_i})$, where J is the cross-entropy loss function and $\hat{y_i} = f(x_i)$. $f(x_i)$ is the output of the neural network function, and three routes have different $f(x_i)$ output.

Let $F$ be the combination function of $L_1$, $L_2$, and $L_3$, and it mainly controls the weights of these losses. Our loss function is depicted in the equation (\ref{eq:general}):

\begin{equation}
    L = F(L_1, L_2, L_3)
    \label{eq:general}
\end{equation}

During the transfer stage, we would like to ensure that the master network effectively learns new features from the target domain without excessively forgetting the pre-trained knowledge from the source dataset. We carried out the mini-batch for training the model. For each training batch, the final loss function is $F = \alpha (L_1 + L_2) + L_3$, where $\alpha$ is a dynamic value to adjust the importance of two loss functions over the $F$. It is used to prevent $L_1$ and $L_2$ from becoming excessively large and potentially interfering with the network's capacity to adapt to the information from the target domain during transfer learning. Meanwhile, it upholds the detection performance of the original domain ($L_3$) and mitigates the risk of catastrophic forgetting. Denoting the neural network's training weights as $\omega$, the gradient expression for the back-propagation about $F$ is depicted in equation (\ref{eq:updated}). 

\begin{equation}
\label{eq:updated}
\frac{\partial{F}}{\partial{\omega}} = \alpha \cdot \left(\frac{\partial{L_1}}{\partial{\omega}} + \frac{\partial{L_2}}{\partial{\omega}}\right) + \frac{\partial{L_3}}{\partial{\omega}}
\end{equation}

We adopted the suggestion by Song et al \cite{song2021robust}, the parameter $\alpha$ for each mini-batch $j$ can be calculated by the equation (\ref{eq:factor}),
\begin{equation}
    \label{eq:factor}
    \alpha = \frac{2L^j_3}{L^j_1 + L^j_2},
\end{equation}
where $L^j_1$, $L^j_2$, and $L^j_3$ are the $L_1$, $L_2$, and $L_3$ losses at the $j$-th mini-batch, respectively. $\alpha$ is initialized with 0. 
We provide the pseudo-code for this part in Alg. \ref{alg:1} from lines 1 to 14.

\begin{algorithm}[t]
\caption{X-Transfer Algorithm}
\textbf{Input}: $\{(x_i, y_i)\}_{i=1}^N$ with data size $N$, a dataset of real and generated images and their labels \\
\textbf{Input}: Pre-trained model: auxiliary network \textbf{f} and master network \textbf{g} \\
\textbf{Hyperparameters}: $s > 0$, $\beta \in (0, 1)$,  $B > 0$
\begin{algorithmic}[1]
    \State \textbf{X-Net}: the composition of sibling network for X-Transfer
    \State $m \in \{0, 1\}$ is a constant \Comment{Control the entrance of \textbf{X-Net}}
    \For{$k \in \{1, 2, ..., \frac{N}{B}\}$} \Comment{Batch iterations}
        \State $\{L_1, L_2, L_3, L_{AUC}, \alpha\} = 0$ \Comment{Initialize sums}
        \State  $\{X, Y\} = \{(x_j, y_j)\}_{j=1}^B$ \Comment{Get next batch}  
        \State $Out_1$ = \textbf{X-Net}($\{X|m=0\}$) \Comment{forward loss start from auxiliary net}
        \State $Out_2$ = \textbf{X-Net}($\{X|m=1\}$) \Comment{forward loss start from master net}
        \State $Out_3$ = \textbf{g}(X) \Comment{output of master network}
        \State $\{L_1, L_2\}$ = CrossEntropy($\{Out_1, Out_2\}$, $\{Y\}$)
        \State $L_3$ = CrossEntropy($\{Out_3\}$, $\{Y\}$)
        \State $\alpha$ = $2 \cdot L_3 / (L_1 + L_2)$
        \State $L_{AUC}$ = AUC($\{Out_3\}$, $\{Y\}$) \Comment{AUC Loss}
        \State $Loss_{main}$ = $\beta \cdot L_3 + (1 - \beta) \cdot L_{AUC}$
        \State \textbf{Target Loss} = $\alpha \cdot (L_1 + L_2) + Loss_{main} + s \cdot \Omega_{fc}(\omega)$
        \State Update the master network g and minimize the Target Loss for 
this batch.
    \EndFor
\end{algorithmic}
\label{alg:1}
\end{algorithm}

\subsection{AUC For Improving Imbalanced Learning}

In machine learning, common evaluation criteria for classification results include accuracy, F1 score, recall, and AUC. While most of these metrics generally yield satisfactory performance on balanced datasets, their effectiveness may diminish in the presence of imbalanced datasets. Accuracy becomes less favorable due to the need for a predefined threshold. Compared to accuracy, F1, recall, AUC, and AP emerge as superior metric choices when dealing with imbalanced datasets, as they obviate the requirement for a threshold. In this study, we opt for AUC as our evaluation metric, with the specific objective of directly maximizing the AUC metric to uphold the model's performance.

AUC metric can measure a model’s performance on a binary classification problem. Nevertheless, directly incorporating the AUC metric into our loss function is challenging due to its nondifferentiability. To address this, we leverage an alternative target loss term to maximize the AUC metric: the normalized Wilcoxon-Mann-Whitney (WMW) statistic. This statistic effectively simulates AUC and possesses differentiability, making it amenable for integration into our loss function.

Given a labeled dataset $\{(x_k, y_k)\}^M_{m=1}$ with the size of $M$ samples, $x$ is the input data and $y$ is its label ($y \in \{0, 1\}$). Then we split the dataset into two different sub-sets as P = $\{k| y_k = 1\}$, and N = $\{k| y_k = 0\}$ where P is the set of positive samples, and N is the set of negative samples. Let $\mathcal{F}: \mathbb{R}^d \rightarrow \mathbb{R}$, $\mathcal{F}$ is a predict function that maps a $d$-dimensional tensor to a one-dimensional number. $\mathcal{F}(x_k)$ represents the predication score of sample $x_k$, where $k \in [1, M]$. 

According to Wilcoxon Mann Whitney statistics ({WMW}) \cite{yan2003optimizing}, we can get an approximate AUC loss as follows:
\begin{equation}
\label{eq:wmw}
    L_{AUC} = \frac{1}{|P||N|} \sum_{i \in P} \sum_{j \in N}
    R(\mathcal{F}(x_i), \mathcal{F}(x_j))
\end{equation}

\begin{equation}
\fontsize{8.5pt}{8pt}
R(\mathcal{F}(x_i), \mathcal{F}(x_j)) = \left\{
\begin{aligned}
& (- (\mathcal{F}(x_i) - \mathcal{F}(x_j) - \gamma))^p, otherwise \\
& 0, \mathcal{F}(x_i) - \mathcal{F}(x_j) \geq \gamma \\
\end{aligned}
\right.
\end{equation}
where $\gamma \in (0, 1]$ and $p > 1$ are two hyper-parameters. Based on our preliminary experiments, we set $\gamma$ as 0.16 and p as 2.0.

\subsection{Overall Loss}
Finally, we define a hyper-parameter $\beta \in [0, 1]$ to balance the three binary cross-entropy (BCE) loss functions ($L_1, L_2, L_3$) and AUC loss function $L_{AUC}$, the overall equation is shown as below:
\begin{equation}
\label{lb:target}
L = \alpha (L_1 + L_2) + \beta L_3 + (1 - \beta) L_{AUC} + s \cdot \Omega_{fc}(\omega)
\end{equation}
where $\Omega_{fc}$ is the $L2$ regulation function, and $\omega$ is the parameters of its linear layer. Meanwhile, s is a hyper-parameter of the $L2$ regulation function, which controls normalization strength.
We will analyze the influences of $\beta$ in the Section \textit{Ablation Study}. And we provide the pseudo-code in Alg.\ref{alg:1} from lines 12 to 14.

\section{Experiments \& Results}
\label{sc:ex}
\subsection{Experimental Settings}

\textbf{Datasets}. In our experiment, we used four distinct datasets, which are publicly available, to evaluate our method. The summary of these datasets is presented in Table.\ref{tab:ds_info}. Each dataset comprises two types of images, real and fake, where the deep learning approaches generate the latter. For instance, the real images of {StarGAN} \cite{choi2018stargan} are provided by {CelebA} \cite{liu2015faceattributes} while the StarGAN model develops the fake images. The real images of both StyleGAN and StyleGAN2 \cite{karras2019style, karras2020analyzing} are created by FFHQ \cite{karras2019style}, and the fake images of them are generated by the StyleGAN and StyleGAN2, respectively (Fig.~\ref{fig:fake_imgs}). The real image of ProGAN classes is LSUN\cite{2015LSUN}, and ProGAN generates its corresponding fake images. The datasets are all balanced, which means the number of real images and the number of fake images are equal. We separated the dataset for the training and testing. The row in the table shows the data size in our experiment. In addition, we isolated 2000 images for each dataset to carry out the transfer learning process. We evaluate the impact of AUC Loss on the imbalanced dataset in the ablation study section \ref{subsc:alhpa}.

\textbf{Baselines}.
We compare our method X-Transfer with different existing approaches, General-Transfer and T-GD, which will be explained as follows. Prior to the comparison, we identified the optimized hyper-parameter settings and chose the best practice based on our preliminary experiments. For instance, we conducted various experiments, including analyzing the effect of $\beta$, investigating the data augmentation, and generalizing our model on the non-face image dataset and performance on an imbalanced dataset. The ablation study will describe the details (see section \ref{subsc:alhpa}). 

\begin{itemize}
    \item \textit{General-Transfer}.
    The typical way to perform transfer learning is by freezing some weights of a pre-trained model learned from the source dataset and fine-tuning it with weight decay for the target dataset. We achieved this method for GAN image detection and called it General-Transfer. To clarify, we froze most of the convolutional layers except for the final convolutional block and linear layers.  The backbone model is RepVGG-A0, and General-Transfer was trained for 300 epochs in the transfer stage with a learning rate of 0.04 and a momentum rate of 0.1. In addition, we employed different data augmentation methods such as RandomHorizontalFlip, JPEG compression and Gaussian blur. At the same time, we also used stochastic gradient descent (SGD) and cosine annealing algorithm \cite{loshchilov2016sgdr} to adjust learning rate decay dynamically. The training step can be terminated in advance when the learning rate drops to 0, or the metric does not change greatly.
    
    \item \textit{T-GD}.
    In addition to general transfer, we conducted T-GD as our second baseline. This method is a SOTA model and original from Jeon et al. \cite{jeon2020t}. We modified it by 300 epochs in the pre-train stage and over 300 epochs in the transfer stage. In addition, we use CutMix \cite{yun2019cutmix} for data augmentation to improve the performance in detecting GAN images during the transfer stage. T-GD uses L2 regularization to constrain the variation of the convolutional layer parameters, thereby preserving the source dataset's performance and preventing catastrophic forgetting. We use the RepVGG-A0 network \cite{ding2021repvgg} as the binary classifier of our backbone because of its high accuracy and computational efficiency. We followed the same data augmentation as General-Transfer and T-GD. All the probability of data augmentation is $50\%$. 
\end{itemize}

For our \textbf{X-Transfer} model, we employ the same experimental metric AUC and dataset as T-GD. We set the initial learning rate as 0.002 in the transfer stage, using low momentum by 0.001 for stochastic gradient descent ({SGD}) inspired by \cite{2020Rethinking}. In addition, we have set JPEG data augmentation with random compression quality, which samples randomly from 30 to 100. We use the Area under the Receiver Operating Characterization Curve(AUC) to evaluate the general performance of the model. Since the accuracy metric requires a fixed threshold to compute the effect of classification in advance, the AUC metric standard is more applicable than it is.

\begin{table}[t]
\centering
\resizebox{1.0 \columnwidth}{!} {
    \begin{tabular}{c|c|c|c|c}
    \toprule
    \textbf{Dataset} & \textbf{StarGAN} & \textbf{StyleGAN} & \textbf{StyleGAN2} & \textbf{ProGAN} \\
    \midrule
    Resolution & 128 × 128 & 256 × 256 & 256 × 256 & 256 × 256 \\
    Train & 137,239 & 33,739 & 42,356 & 288,000 \\
    Test & 50,000 & 30,000 & 30,000 & 32,000 \\
    Transfer & 2000 & 2000 & 2000 & 2000 \\
    \bottomrule
    \end{tabular}
}
\caption{An overview of datasets}
\label{tab:ds_info}
\end{table}


\subsection{Comparison with Baseline}
Our experimental results are depicted in Table.\ref{tab:ex_results}. We trained the model on the source dataset and conducted transfer learning on the target dataset. Specifically, two datasets are used in the comparison: the source dataset is StarGAN, and transfer learning on StyleGAN2 or vice versa. The AUC is calculated on the testing dataset. We compared our X-Transfer model against the General-Transfer and T-GD, respectively, and then we added the AUC loss in our model to improve the performance.

\textbf{General-Transfer \textit{vs} X-Transfer}.
In the transfer stage, the General-Transfer method freezes the weights of the pre-trained model and fine-tunes it to learn from the target dataset. During this period, the General-Transfer presents a trade-off between the performance of the source and the target datasets.  Our experimental results are depicted in Table.\ref{tab:ex_results}. After transfer learning, we observed that General-Transfer achieved promising results on the first task (from StarGAN to StyleGAN2), where the AUC on both source and target datasets is over 91\%. However, the results of the target dataset in the second transfer learning task (from StyleGAN2 to StarGAN) decreased dramatically to 59.8\%, which is spuriously lower compared to the source dataset (99.40\%).


On the contrary, X-Transfer shows more consistent performance on both the source and target datasets. It maintains the high performance (above 99\% AUC) achieved on the source dataset while transferring to the target dataset, resulting in impressive scores on both datasets.

\textbf{T-GD \textit{vs} X-Transfer}.
T-GD adopts an L2 regularization to limit the variation of layers' parameters, which can maintain the performance on both the source and target datasets. Meanwhile, T-GD utilizes an adaptive iterative training method based on the two alternative networks, which can enhance the model's ability to acquire knowledge better.

X-Transfer uses a sibling network to transport gradients in the order of corresponding blocks. Regarding experimental results, X-Transfer performs better on both the source and target datasets. When the model was trained by StyleGAN2 and transferred to StarGAN, X-Transfer exhibited an advanced AUC metric(99.90\%) on the source dataset and target dataset. In contrast, T-GD's metric(90.01\% and 90.0\%) was lower than X-Transfer, no matter the source or target dataset. 

\begin{table}[t]
    \centering
    \resizebox{1.0 \columnwidth}{!} {
        \begin{tabular}{c|c|c|c|c}
        \toprule
            \textbf{\makecell{Source \\ dataset}} & \textbf{\makecell{Target \\ dataset}} & \textbf{Methods} & 
            \textbf{\makecell{Source \\ AUC}} & \textbf{\makecell{Target \\ AUC}} \\
            \midrule
            \multirow{4}{*}{StarGAN} & \multirow{4}{*}{StyleGAN2} & General-Transfer & 91.09\% & 95.70\% \\
            \multirow{4}{*}{} & \multirow{4}{*}{} & T-GD & 96.4\% & 90.1\% \\
            \multirow{4}{*}{} & \multirow{4}{*}{} & X-Transfer & 98.20\% & 90.21\% \\
            \multirow{4}{*}{} & \multirow{4}{*}{} & X-Transfer + AUC Loss & \textbf{99.04\%} & 90.12\% \\
            \cmidrule(lr){1-5}
            \multirow{4}{*}{StyleGAN2} & \multirow{4}{*}{StarGAN} & General-Transfer & 99.40\% & 59.80\% \\
            \multirow{4}{*}{} & \multirow{4}{*}{} & T-GD & 90.01\% & 90.0\% \\
            \multirow{4}{*}{} & \multirow{4}{*}{} & X-Transfer & 99.72\% & 92.30\% \\
            \multirow{4}{*}{} & \multirow{4}{*}{} & X-Transfer + AUC Loss & \textbf{99.90\%} & \textbf{94.54\%} \\
            \bottomrule
        \end{tabular}
    }
    \caption{The results of transfer learning using AUC metric}
    \label{tab:ex_results}
\end{table}

Moreover, X-Transfer uses fewer epochs than T-GD in transfer learning. In the transfer stage, T-GD spent at least 300 epochs, but X-Transfer only spent 10\% of its epochs on average, both of which were trained with the same dataset. 

\textbf{With AUC Loss \textit{vs} Without AUC Loss}.
We compare the difference between our method with and without the AUC Loss function when using the X-Transfer method.  The primary target function combines binary cross-entropy, L2 regularization, and the AUC Loss function. As illustrated in Table.\ref{tab:ex_results} above, irrespective of the dataset employed, utilizing AUC Loss consistently leads to improved performance, with at least one of the source or target datasets surpassing the performance achieved without AUC Loss. For example, the use of AUC on X-Transfer is 0.18\% and 2.24\% higher than the model without AUC on both datasets(when transferring from StyleGAN2 to StarGAN). The experimental results also show that AUC Loss positively affects the transfer stage, regardless of whether it is applied to the source or target dataset.

\subsection{Comparison with non-transfer learning methods}
We compare our method with recently excellent non-transfer learning-based methods to verify the generalization and robustness of our method. We use StarGAN, CycleGAN, BigGAN\cite{brock2018large}, GauGAN\cite{park2019semantic}, StyleGAN and StyleGAN2 from ~\cite{wang2020cnn} as the test set to evaluate its effects in Table.\ref{tab:cross_models}. 
Among them, the real datasets of ProGAN, StyleGAN, and StyleGAN2 are composed of LSUN; BigGAN and GauGAN are composed of ImageNet and COCO datasets, respectively. The corresponding models generate all fake samples.
After all, we also compare with some popular published studies~\cite{jeong2022frepgan,wang2020cnn,gragnaniello2021gan,tan2023learning} and general CNN architectures~\cite{tan2019efficientnet,chollet2017xception,liu2022convnet}. All the evaluation metrics are based on AP (Average Precision).

Experiments have proven that our method performs best on CycleGAN, BigGAN dataset, and average, which illustrates the significant performance of our method.

\begin{table*}[h]
  \centering
  \resizebox{1.9\columnwidth}{!} {
      \begin{tabular}{c|c|c|c|c|c|c|c|c}
      \toprule
          \textbf{Methods} & \textbf{ProGAN} & \textbf{StarGAN} & \textbf{StyleGAN} & \textbf{CycleGAN} & \textbf{BigGAN} & \textbf{StyleGAN2} & \textbf{GauGAN} & \textbf{Mean} \\
          \midrule
          EfficientNet-B0~\cite{tan2019efficientnet} & 100 & 99.9 & 99.9 & 95.9 & 87.5 & 99.8 & 83.3 & 95.2 \\
          ConvNext-Tiny~\cite{liu2022convnet} & 100 & 100 & 99.8 & 96 & 82.3 & 99.9 & 80.1 & 94.0 \\ 
          Xception~\cite{chollet2017xception} & 100 & 99.2 & 99.6 & 95.6 & 86.0 & 99.8 & 77.1 & 93.9 \\
          FrePGAN~\cite{jeong2022frepgan} & 99.4 & 100 & 90.6 & 59.9 & 60.5 & 93.0 & 49.1 & 78.9 \\
          Wang~\cite{wang2020cnn} & 100 & 95.4 & 98.5 & 96.8 & 88.2& 99.1 & 98.1 & 96.6 \\
          Tan~\cite{tan2023learning} & 99.9 & 100 & 99.6 & 94.4 & 88.9 & 99.4 & 81.8 & 94.9 \\
          Gragnaniello~\cite{gragnaniello2021gan} & 100 & 97.1 & 98.3 & 82.2 & 78.1 & 99.2 & 85.6 & 94.9 \\
          \textbf{Ours} & 100 & 97.7 & 98.0 & \textbf{98.7} & \textbf{91.8} & 98.7 & 92.5 & \textbf{96.8} \\
          \bottomrule
      \end{tabular}
      }        
  \caption{Results of cross GAN models}
  \label{tab:cross_models}
\end{table*}

\subsection{Ablation Study}
\label{subsc:alhpa}
\begin{figure}[ht]
    \centering
    \includegraphics[width=1.0\linewidth]{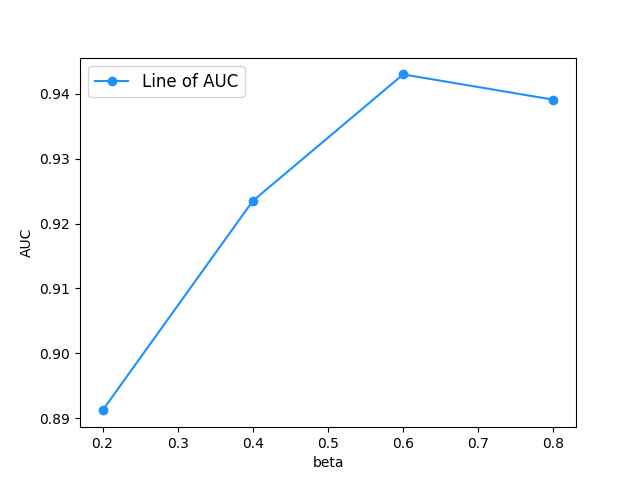}
    \caption{Impact of hyper-parameter $\beta$ in AUC Loss in target dataset.}
    \label{fig:betas}
\end{figure}

To identify the best practice settings for our comparison, we carried out several ablation experiments as follows:

\textbf{Hyper-parameter Analysis for AUC}.
The selection of hyper-parameter $\beta \in [0, 1]$ can balance the binary cross-entropy(BCE) loss function and AUC loss function $L_{AUC}$ as shown in Equation \ref{lb:target}. We used different hyper-parameter $\beta$ to evaluate the effect of AUC Loss. We trained our model with different $\beta$ values. According to the results displayed in Figure.\ref{fig:betas}, $\beta$ = 0.6 yields the best detection performance while transferring StyleGAN2 to StarGAN. 

\textbf{Effect of Data Augmentation}.
Apart from the factors discussed above, we also investigated the impact of data augmentation in our experiment. We applied various data augmentation methods, such as Gaussian blur, JPEG compression, and CutMix~\cite{yun2019cutmix}. 

As depicted in Table.\ref{tab:eff_aug}, we investigated the impact of data augmentation on our X-Transfer method. We used StarGAN as the source dataset and transferred it to StyleGAN as the target dataset. After applying data augmentation, the results show a notable increase of 5.6\% in the AUC score on the source dataset, albeit with a slight decrease in the AUC on the target dataset, moving from 98.9\% to 97.4\%. In this context, we recommend employing augmentation to augment the dataset size and mitigate the risk of overfitting during the transfer stage.


\begin{table}[H]
    \centering
    \resizebox{1.0 \columnwidth}{!} {
        \begin{tabular}{c|c|c|c|c}
        \toprule
            \textbf{Method} & \textbf{\makecell{Source dataset}} & \textbf{\makecell{Target dataset}} & \textbf{\makecell{Source \\ AUC}} & \textbf{\makecell{Target \\ AUC}}  \\
            \midrule
            w/o augmentaion & \multirow{2}{*}{StarGAN} & \multirow{2}{*}{StyleGAN} & 93.2\% & 98.9\% \\
            w augmentation & \multirow{2}{*}{} & \multirow{2}{*}{} & 98.8\% & 97.4\% \\
            \bottomrule
        \end{tabular}
    }
    \caption{Impact of data augmentation in X-Transfer}    
    \label{tab:eff_aug}
\end{table}

\textbf{Effect on Non-face Dataset}.
Instead of limiting our method to facial image detection, we assessed X-Transfer on non-face datasets as well, employing three distinct types of datasets as the target dataset. The non-face datasets include LSUN-Cat, LSUN-Church, and LSUN-Car. All the real datasets were obtained from the sources mentioned earlier (real ones from the CelebA and LSUN datasets and fake ones from StarGAN and StyleGAN2 generated from real ones). The source dataset was StarGAN, and the target dataset was a combination of the abovementioned datasets and StyleGAN2.

\begin{table}[H]
    \centering

    \resizebox{0.65 \columnwidth}{!} {
        \begin{tabular}{c|c|c|c}
            \toprule
            \textbf{Test Dataset} & \textbf{Cat} & \textbf{Church} & \textbf{Horse} \\ \midrule
            AUC & 97.9\% & 98.1\% & 97.6\% \\ \bottomrule
        \end{tabular}
    }
    \label{tab:non-face}    
    \caption{Impact on non-face dataset}
\end{table}

All results in Table.~\ref{tab:non-face} are over 97\%, and the best performance is using the LSUN-Church dataset, proving that our method can detect non-facial images effectively.

\textbf{Effect of AUC Loss on imbalanced dataset}
To evaluate the effect of AUC loss on the imbalanced dataset, we constructed a dataset of GauGAN with a real-to-fake ratio close to 5:1. We also verified the impact on the final detection performance without using AUC Loss and using it. After experimental validation, using AUC Loss has better detection performance than not using it on the imbalanced dataset.

\begin{table}[H]
    \centering

    \resizebox{1.0 \columnwidth}{!} {
    \begin{tabular}{c|c|c|c|c} \toprule
        \textbf{Method} & \textbf{\makecell{Source Dataset}} & \textbf{\makecell{Target Dataset}} & \textbf{\makecell{Source \\ AUC}} & \textbf{\makecell{Target \\ AUC}}  \\ \midrule
        w/o AUC Loss & \multirow{2}{*}{ProGAN} & \multirow{2}{*}{GauGAN} & 99.5\% & 87.2\% \\
        w AUC Loss & \multirow{2}{*}{} & \multirow{2}{*}{} & 99.1\% & 91.2\% \\ \bottomrule
    \end{tabular}
    }    
    \caption{Impact of AUC Loss on imbalanced dataset}
    \label{tab:eff_auc_loss}
\end{table}

The outcome is presented in Table.~\ref{tab:eff_auc_loss} indicates a 4\% increase in AUC on the target dataset when AUC Loss is applied, demonstrating that this feature enhances the overall detection performance, irrespective of dataset balance.

\section{Conclusion}
\label{sc:conc}
This paper presents X-Transfer, an innovative detection algorithm designed for identifying GAN-generated images. It features two identical networks—a source-retaining auxiliary network and a target-absorbing master network. In this setup, gradients propagate alternatively through corresponding layers, outperforming conventional transfer approaches and achieving a success rate of 99.04\% in our experiments. Additionally, we introduce AUC Loss, ensuring a higher AUC metric and proving beneficial for training a high-performance detection model. Lastly, we conduct a series of comprehensive experiments demonstrating the promising results of X-Transfer across different hyper-parameter settings, data augmentation techniques, model generalization approaches, and imbalanced datasets.

In the future, recognizing that GAN-generated images encompass various types of fake images, we aspire to broaden the application of this approach to diverse datasets beyond facial images.

\bibliography{maindraft}

\end{document}